\documentclass[conference]{IEEEtran}

\usepackage{cite}
\usepackage{float}
\usepackage{booktabs}
\usepackage{stfloats}
\usepackage{url} 
\usepackage{tabularx,booktabs}
\usepackage{amsmath,amssymb,amsfonts}
\usepackage{algorithmic}
\usepackage{graphicx}
\usepackage{textcomp}
\usepackage{xcolor}

\def\BibTeX{{\rm B\kern-.05em{\sc i\kern-.025em b}\kern-.08em
    T\kern-.1667em\lower.7ex\hbox{E}\kern-.125emX}}
\begin{document}

\title{TARA: A Low-Cost 3D-Printed Robotic Arm for Accessible Robotics Education\\

\thanks{Identify applicable funding agency here. If none, delete this.}
}

\author{\IEEEauthorblockN{Thays Leach Mitre}
\IEEEauthorblockA{\textit{Department of Computer Science, Data Science and Engineering} \\
\textit{New York University Shanghai}\\
Shanghai, China \\
tal9567@nyu.edu}

}

\maketitle

\begin{abstract}
       The high cost of robotic platforms limits students’ ability to gain practical skills directly applicable in real-world scenarios. To address this challenge, this paper presents TARA, a low-cost, 3D-printed robotic arm designed for accessible robotics education. TARA includes an open-source repository with design files, assembly instructions, and baseline code, enabling users to build and customize the platform. The system balances affordability and functionality, offering a highly capable robotic arm for approximately \$200, significantly lower than industrial systems that often cost thousands of dollars. Experimental validation confirmed accurate performance in basic manipulation tasks. Rather than focusing on performance benchmarking, this work prioritizes educational reproducibility, providing a platform that students and educators can reliably replicate and extend.
\end{abstract}


\section{Introduction}
Robotics is playing an increasingly vital role in both industry and education. However, the high cost of robotic systems remains a significant obstacle to broader adoption, particularly in educational settings and small-scale industries. This financial barrier limits access to hands-on learning experiences. Prior research has highlighted how low-cost robotics platforms, such as 3D-printed robots and simplified educational kits, can provide students with valuable opportunities to engage in robotics while reducing financial barriers \cite{Armesto2016, Ceccarelli2001}. Early exposure to robotics in K–12 education has also been shown to foster STEM competencies and give students a real-world context for applying knowledge \cite{Cejka2006}. 

\vspace{0.5\baselineskip}
The objective of this paper is to design, assemble, and program a 3D-printed robotic arm that is affordable, easy to build, and capable of performing several robotic tasks. The robot, named TARA, was developed at a total cost of approximately \$200, which is significantly lower than comparable robotic arms, while remaining accessible to students. TARA is designed to perform basic tasks such as picking and placing objects, while also serving as an educational platform that introduces students to robotics design, programming, and hardware integration.
\vspace{0.5\baselineskip}

In addition to the robotic arm itself, the project delivers several contributions. First, it provides organized design files, assembly instructions, and baseline code in an open-source repository, ensuring that others can replicate, customize, and extend the system. Second, a Python-based simulation environment was implemented using Pinocchio and MeshCat, enabling safe testing of robot configurations and collision detection before deployment. Third, two control methods were developed: direct joint angle input for precise positioning, and a potentiometer-based input for intuitive interaction. Together, these contributions provide a robust and affordable framework for both hands-on and virtual learning.
\vspace{0.5\baselineskip}

The remainder of this paper is organized as follows. Section II reviews related work in low-cost robotics and educational applications. Section III describes the design and implementation of TARA, including its mechanical, electrical, and software components. Section IV presents results and discussion, including validation, cost analysis, and educational impact. Section V concludes with key findings and outlines directions for future work.

\section{Related Work}
\subsection{Educational Robotics}
Constructionist learning theory \cite{Siegler1986}, which emphasizes learning by doing and building tangible artifacts \cite{Papert1991}, has been shown to enhance student engagement and problem-solving skills compared to traditional approaches \cite{Sergeyev2010}. Several educational platforms, such as LEGO Mindstorms, are widely adopted for introducing robotics concepts \cite{Resnick1996}. While these systems successfully foster early engagement, their proprietary and closed nature limits extensibility and prevents students from progressing toward more advanced, real-world robotics applications. To overcome these limitations, researchers have investigated affordable robotic manipulators. For example, Quigley \cite{Quigley2011} developed a compliant 7-DoF manipulator that offered a good balance between performance and usability, but its cost of approximately \$4000 restricts accessibility in educational contexts. Similarly, Sharma \cite{Sharma2012} introduced a nine-DoF anthropomorphic arm capable of advanced manipulation tasks. However, its reliance on EMG-based control increased complexity and introduced signal delays, making it less suitable for novice learners.
\vspace{0.5\baselineskip}

These examples demonstrate that existing educational robotic arms either remain expensive or introduce technical barriers that limit accessibility for beginners. In contrast, TARA provides a low-cost 3D-printed, and open-source alternative that is specifically designed to be both reproducible and customizable. By combining affordability with dual control modes and simulation support, TARA aims to bridge the gap between introductory robotics kits and advanced robotic manipulators, offering a platform that supports both hands-on education and accessible research experimentation. 
\subsection{Low-Cost Industrial Applications} Small and medium-sized enterprises (SMEs) face increasing pressure to improve efficiency and safety while working under strict budget constraints. Although industrial robots can deliver these benefits, their high acquisition and maintenance costs often make them inaccessible to smaller companies. This has created a demand for affordable robotic solutions that provide essential functionality without the expense of full-scale commercial systems.
\vspace{0.5\baselineskip}

Several researchers have explored this direction. Rogers \cite{Rogers2009} developed a low-cost teleoperable arm for hazardous inspection tasks using servos and radio components, demonstrating affordability but with limited precision. Pereira \cite{Pereira2014} introduced a robotic arm for object sorting with Raspberry Pi and GNU Octave, showcasing how open-source software and low-cost hardware can replicate some industrial functions. Elfasakhany \cite{Elfasakhany2011} designed a 4-DoF acrylic-based robotic arm, achieving a balance between performance and cost but restricting durability and scalability. Limeira \cite{Limeira2019} advanced this line with WsBot, a low-cost swarm robot aimed at Industry 4.0 experimentation, though its miniature scale limited real-world adoption. More recently, Ali \cite{Ali2023} proposed a 5-DoF robotic arm fabricated from locally sourced materials and optimized through finite element analysis, highlighting the potential for regionally adapted low-cost solutions.
\vspace{0.5\baselineskip}

While these approaches demonstrate the feasibility of affordable robotics, most rely on metals or acrylics that require machining, which increases expense and reduces accessibility for institutions without advanced manufacturing tools. In contrast, TARA employs 3D-printed components that are less expensive, safe to fabricate, and easily customizable.
\vspace{0.5\baselineskip}

By openly publishing design files, TARA enables SMEs and educational institutions to produce parts in-house with minimal resources, significantly lowering barriers to adoption. The trade-offs between different material choices for robotic arms are summarized in Table~\ref{tab:tab1}. Although not intended for heavy-duty industrial lifting, TARA’s use of 3D-printed plastics provides a practical compromise between cost, usability, and safety, making it a suitable platform for both educational contexts and small-scale industrial applications.

\begin{table}[htbp]
\caption{Comparison of materials used to develop robotic arms}
\label{tab:tab1}
\centering
\scriptsize
\renewcommand{\arraystretch}{1.2}
\begin{tabularx}{\columnwidth}{@{}l l X X@{}}
\toprule
\textbf{Material} & \textbf{Cost} & \textbf{Advantages} & \textbf{Disadvantages} \\
\midrule
Soft Polymers \cite{Georgopoulou2021} & Expensive &
\begin{itemize}
  \item Works in limited space
  \item Motorless deformation mechanism
  \item Flexible deployment
\end{itemize} &
\begin{itemize}
  \item Limited commercial applications
  \item Lifts only lightweight objects
  \item Time-consuming programming
\end{itemize}\\
\midrule
Composites \cite{Chen2019} & Low-cost &
\begin{itemize}
  \item Lightweight
  \item Simple operation
  \item Detects small strains
\end{itemize} &
\begin{itemize}
  \item Delay in response
  \item Hysteresis
\end{itemize}\\
\midrule
Metals and Alloys \cite{Patidar2016} & Cost-effective &
\begin{itemize}
  \item Lift heavy weights
  \item Widely used in production
  \item Embedded security features
\end{itemize} &
\begin{itemize}
  \item Detect only large strains
  \item Heavyweight robot
  \item Expensive if nickel/chromium
  \item Wide workspace required
\end{itemize}\\
\bottomrule
\end{tabularx}
\end{table}

\section{System Design and Implementation}
To address the challenges of creating an affordable, accessible robotic platform, this project followed a structured approach divided into key phases. Each phase focused on a critical aspect of the design, development, and testing of TARA. The timeline below in Figure~\ref{fig:Figure 1} provides an overview of the project’s progression.
\begin{figure*}[h!]
    \centering
    \includegraphics[width=1\linewidth]{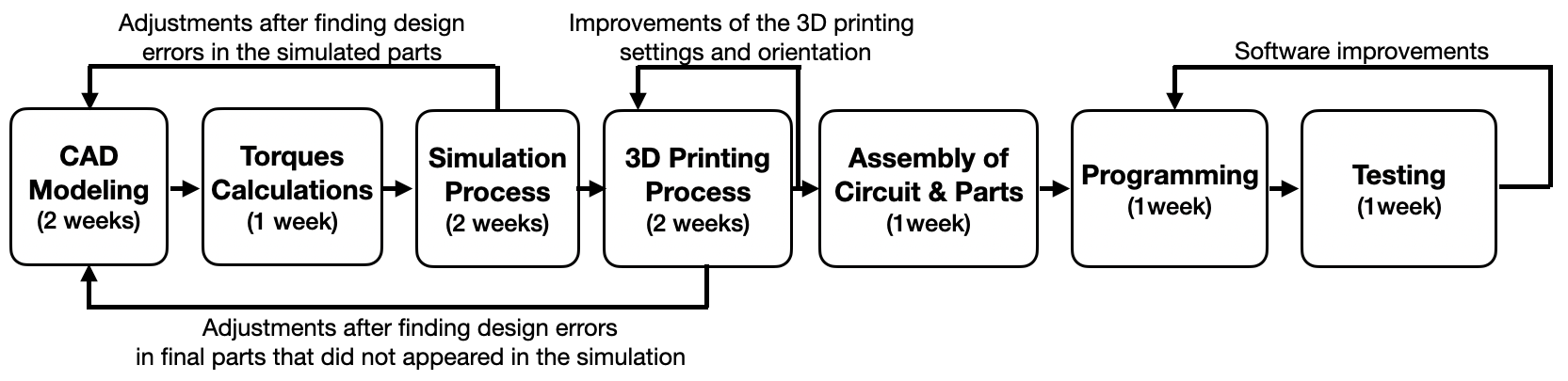}
    \caption{Design, development, and testing workflow of TARA.}
    \label{fig:Figure 1}
\end{figure*}
\subsection{Mechanical Design}
The structural components of TARA were designed in Fusion 360 and exported as stereolithography (\textit{.stl}) files, which describe the surface geometry of 3D objects as triangular meshes. The arm consists of 19 parts, including the base, three primary links, the end-effector, and a scaled-down replica used as a controller. Each link was modeled as two hollow sections to allow integration of motors and cabling.
To determine actuator requirements, the weight of each part was estimated in Fusion 360 by assigning plastic as the design material, chosen for its affordability and ease of fabrication. The required torque at each joint was computed as
\begin{equation}
\tau = \mathbf{r} \times \mathbf{F} \label{eq}
\end{equation}
where \(\tau\) is the torque, F is the gravitational force, and r is the distance from the joint axis to the part’s center of mass. Motors were selected based on these torque estimates to ensure reliable operation.
Parts were fabricated using fused deposition modeling (FDM) 3D printing. Several configurations were tested to balance structural strength with material efficiency. Lightweight plastics reduced torque demands but exhibited bending under self-weight, while rigid plastics provided sufficient stiffness. Final tests used a 20\% infill density and 1.0 mm shell thickness, which offered a compromise between strength and cost.
\subsection{Simulation Environment}
\begin{figure}[htbp]
    \centering
    \includegraphics[width=0.8\columnwidth]{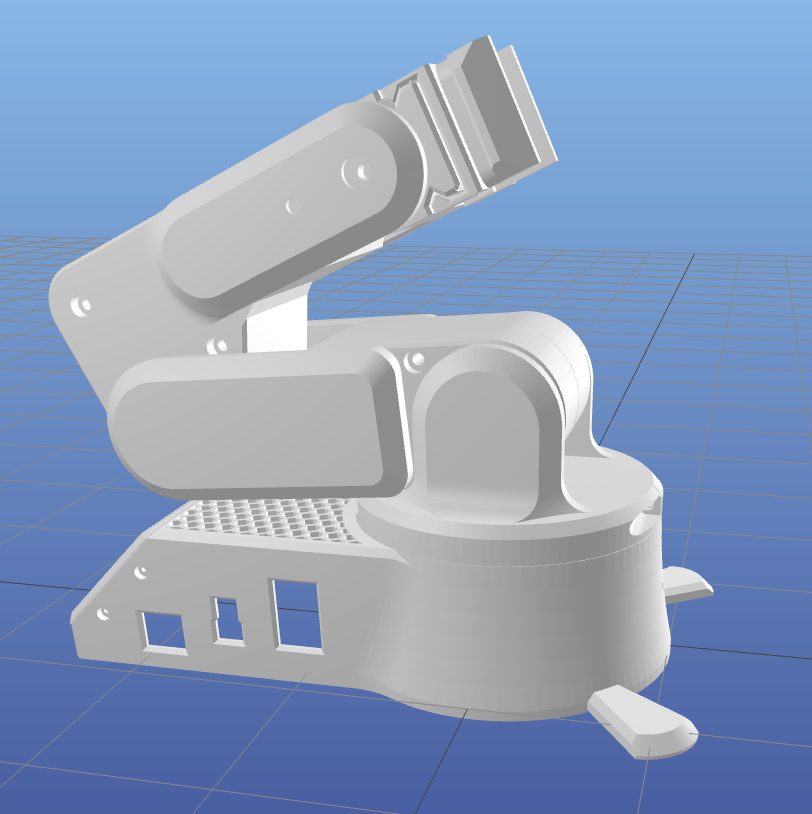}
    \caption{Simulated model of TARA in the Python-based Pinocchio framework with 3D visualization provided by MeshCat. The model illustrates the robot’s body structure and joint configurations used for kinematic and dynamic analysis.}
    \label{fig:simulation}
\end{figure}

To validate the design prior to fabrication, a simulation environment was developed using both CAD models and robotics libraries. The \textit{.stl} files generated in Fusion 360 for 3D printing were reused to reconstruct a virtual representation of the robotic arm, as shown in Figure~\ref{fig:simulation}. This approach enabled preliminary verification of feasibility while reducing reliance on multiple physical prototypes, which is essential in low-cost development.

The physical configuration and kinematic properties of the robot were defined in a Unified Robot Description Format (\textit{URDF}) file, which specified link dimensions, joint types, and axis orientations. This model was integrated into a Jupyter Notebook environment and processed using the \textit{Pinocchio} robotics library in conjunction with the \textit{MeshCat} 3D visualization tool. Pinocchio efficiently computed kinematics and dynamics, including joint positions, velocities, and forces, while MeshCat provided real-time interactive visualization of the robot’s motion in a browser-based interface.
\vspace{0.5\baselineskip}

Two control modes were implemented within the simulation. The first applied forward kinematics, mapping joint input angles to end-effector positions in Cartesian space. The second employed inverse kinematics, using the Jacobian pseudo-inverse method to compute the joint angles required for a desired end-effector position. Iterative updates of the pseudo-inverse reduced position error until it fell below a predefined tolerance or a maximum number of iterations.
\vspace{0.5\baselineskip}

Initial implementations lacked physical constraints, occasionally producing configurations that were not feasible in real hardware, like link collisions. To mitigate this, the \textit{URDF} model was augmented with bounding boxes for each link. This constrained simulation improved physical realism and established a foundation for safe validation of control algorithms prior to deployment on the physical system.

\subsection{Assembly}
The assembly of TARA was carried out in two phases: mechanical construction and electrical integration.
The mechanical assembly began with inspection of the 3D-printed parts to ensure dimensional accuracy and proper alignment. Servo motors were inserted into designated slots within each link’s hollow halves, with gears aligned to the motor shafts to ensure smooth transmission of motion. The two halves of each link were then joined using bolts, providing stability while maintaining the structural integrity of the printed material. Once completed, the links were interconnected through the servo motors to form the complete kinematic chain, ensuring the required range of motion across all joints.
\vspace{0.5\baselineskip}

\begin{figure}[htbp]
    \centering
    \begin{minipage}{0.3\textwidth}
        \centering
        \includegraphics[width=\columnwidth]{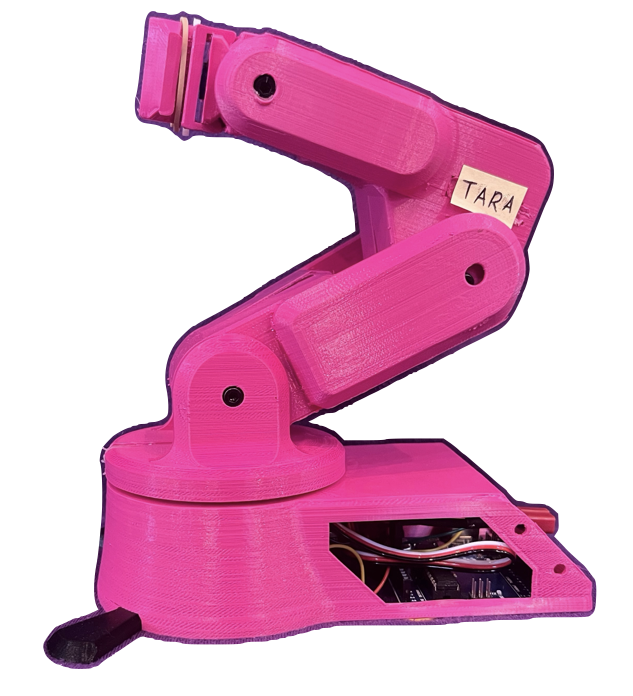}
        \caption{The fully assembled TARA robotic arm, designed using 3D-printed components and servo-based joints.}
        \label{fig:fig12}
    \end{minipage}

    \vspace{1em} 

    \begin{minipage}{0.3\textwidth}
        \centering
        \includegraphics[width=\columnwidth]{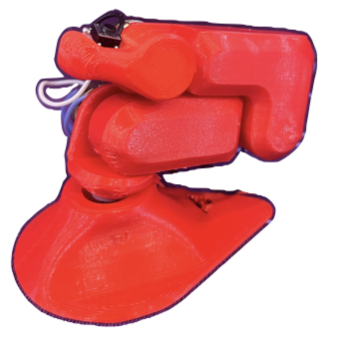}
        \caption{Scaled-down replica of TARA used as a potentiometer-based controller for intuitive teleoperation.}
        \label{fig:fig13}
    \end{minipage}
\end{figure}

The electrical integration involved connecting the servo motors to an Adafruit PWMServoDriver interfaced with an Arduino microcontroller. Potentiometers in the potentiometer-based control system were assigned to each joint: gripper, wrist, elbow, shoulder, and base. The gripper servo was additionally equipped with a push-button actuator for simplified operation. Both the driver board and microcontroller were powered by a regulated supply to provide sufficient current and voltage for the actuators.
Finally, the mechanical and electrical subsystems were combined into a single platform. Wiring was routed internally through the arm to reduce interference with moving components. The completed robotic arm is shown in Figures~\ref{fig:fig12} and~\ref{fig:fig13}.

\subsection{Control and Programming}
The control architecture of TARA was implemented using two modes to support both precise testing and intuitive operation.
The first mode, \textit{Direct Angle Input}, allows user-defined joint angles to be mapped directly to servo motor positions via the Adafruit PWMServoDriver library. This approach enables precise positioning of the robotic arm and is particularly suited for fixed trajectories and validation experiments. 
The second mode, \textit{Potentiometer-Based Input}, translates analog voltage readings from potentiometers into joint angles through a linear mapping function. This allows real-time manual control of the robotic arm. A push-button was integrated to actuate the gripper.
\vspace{0.5\baselineskip}

To ensure safe operation, a validation protocol was incorporated into the direct input mode. Before commands are executed on the physical robot, the same joint angles are tested in the simulation environment to detect potential collisions between links. If no collisions are detected, the motion is executed; otherwise, a warning is issued. By contrast, the potentiometer-based mode inherently respects the physical constraints of the hardware, eliminating the need for pre-simulation validation.

\section{Results and Discussion}

\subsection{System Performance}
TARA achieved its design objectives as a functional, low-cost robotic arm. The platform successfully performed fundamental tasks such as reaching specified positions and executing commands through the two control modes. Simulation outputs were consistent with physical tests, validating the accuracy of the design.
\vspace{0.5\baselineskip}

\begin{figure}[htbp]
    \centering
    \includegraphics[width=\columnwidth]{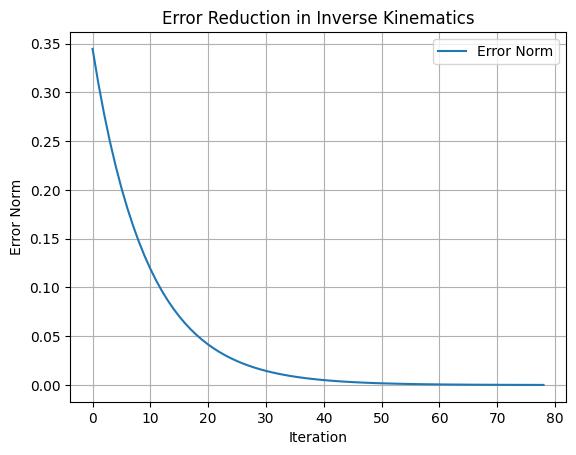}
    \caption{Error reduction in the inverse kinematics solver. 
    The error norm between the desired and actual end-effector position decreases rapidly with successive iterations, 
    demonstrating stable convergence using the Jacobian pseudo-inverse approach.}
    \label{fig:graph}
\end{figure}

The inverse kinematics solver demonstrated stable convergence, as shown in Figure~\ref{fig:graph}. The error norm between desired and actual end-effector positions decreased rapidly within 20 iterations, confirming the feasibility of the Jacobian pseudo-inverse approach for real-time applications.

\subsection{Educational and Practical Impact}
The design and implementation process emphasized skill development in CAD modeling, Python, C++, and hardware-software integration. The platform provides both hands-on and virtual learning opportunities by combining a physical arm with a simulation environment, which aligns with educational goals for robotics training.

\subsection{Limitations}
Although TARA achieved its design objectives, several constraints remain evident from the evaluation. The structural durability is limited by the use of 3D-printed plastic, restricting payload capacity and long-term robustness compared to metallic arms. The reduced number of degrees of freedom narrows the range of tasks relative to industrial manipulators. Additionally, the current separation between the simulation environment (Python/Pinocchio) and the control system (Arduino) prevents integration and real-time hardware-in-the-loop testing.
\begin{table*}[!b]
\centering
\caption{Bill of materials and total estimated cost of TARA}
\label{tab:tab2}
\begin{tabular}{l l c c}
\toprule
\textbf{Material} & \textbf{Description} & \textbf{Quantity} & \textbf{Cost (USD)} \\
\midrule
Standard Servomotor      & Allows the joints to move to specific positions & 1 & 13.98 \\
High Torque Servomotor   & For the shoulder joint that handles extra weight & 3 & 17.99 \\
Micro Servomotor         & For the gripper where less force is required & 1 & 7.99 \\
Arduino Uno Board        & Microcontroller to control the system & 1 & 27.60 \\
Servo Driver Module      & To control the servomotors & 1 & 13.99 \\
Potentiometers           & Sensors for controller joints & 4 & 6.99 \\
Tactile Push Button      & To close the gripper & 1 & 7.56 \\
Power Switch             & To turn the robot ON/OFF & 1 & 6.99 \\
Adjustable Power Supply (7V) & Provides sufficient power to motors & 1 & 25.99 \\
T Plug Connector         & Connects power supply to the system & 1 & 8.97 \\
Wire and Connector Set   & Connects motors and sensors to boards & 1 & 14.99 \\
Gripper Gears            & Enables gripper joint motion & 1 & 9.79 \\
Filament Roll            & Material for 3D printing & 1 & 15.99 \\
Screw Set                & Fits the links together & 1 & 8.99 \\
Long Wire Set            & Connects potentiometers & 1 & 9.99 \\
\midrule
\textbf{TOTAL} & & & \textbf{197.80} \\
\bottomrule
\end{tabular}
\end{table*}
\subsection{Cost Analysis}
Table~\ref{tab:tab2} summarizes the bill of materials. The total cost of the system was \$197.80, significantly lower than comparable robotic platforms that often exceed several thousand dollars.
\vspace{0.5\baselineskip}

\section{Conclusion and Future Work}
This work presented the design and development of TARA, a low-cost, 3D-printed robotic arm aimed at democratizing access to robotics education. With a total cost of approximately \$200, significantly lower than comparable systems. TARA provides both simulation-based and hands-on learning opportunities. The platform integrates two control modes (direct joint angle input and a potentiometer-based interface) and a Python-based simulation environment built on Pinocchio and MeshCat. To promote reproducibility and further adoption, all design files, assembly instructions, and control code are openly available in a GitHub repository:
\url{https://github.com/Thayslm30/Three-D-Printed-Affordable-Robotic-Arm/tree/main}.
In addition, demonstration videos showcasing TARA’s capabilities are provided online:
\url{https://drive.google.com/drive/folders/1L06COeyFpNktWuyYnfT8AG40fpObh4bi?usp=sharing}.
\vspace{0.5\baselineskip}

Experimental validation showed close agreement between simulation and physical results, confirming design accuracy and safety. The arm successfully performed fundamental tasks such as joint-specific positioning and small object manipulation, establishing its value as an accessible and hands-on educational platform.
\vspace{0.5\baselineskip}

Looking ahead, the observed limitations naturally suggest future directions. Structural durability may be improved through hybrid materials or reinforced 3D-printing strategies. Increasing the arm’s degrees of freedom would broaden task capabilities, while unifying the simulation and control systems through softwares like ROS, could enable real-time hardware-in-the-loop testing. Wireless communication (Bluetooth/Wi-Fi) and advanced sensors such as encoders, IMUs, or cameras could further enhance precision and flexibility. Ultimately, TARA is envisioned not only as an introductory educational tool but also as a scalable, customizable testbed for students and researchers exploring advanced robotics concepts.

\section*{Acknowledgment}
The author would like to thank Prof. Ludovic Righetti for his guidance and support as advisor, whose encouragement and provision of resources made this project possible. Appreciation is also extended to Prof. Prométhée Spathis for his valuable advice on improving the clarity of both written and oral presentations. The author further acknowledges Prof. Nasir Memon for his support throughout the course of this work.

\end{document}